\documentclass{article}

\PassOptionsToPackage{numbers,compress}{natbib}
\usepackage[preprint]{neurips_2026}

\usepackage[utf8]{inputenc}
\usepackage[T1]{fontenc}
\usepackage[pagebackref=true,breaklinks,colorlinks,citecolor=blue,linkcolor=blue,urlcolor=blue,hypertexnames=false]{hyperref}
\usepackage{url}
\usepackage{booktabs}
\usepackage{amsfonts}
\usepackage{amsmath}
\usepackage{nicefrac}

\usepackage{microtype}
\usepackage[table]{xcolor}
\usepackage{graphicx}
\usepackage{subcaption}
\usepackage{multirow}
\usepackage{wrapfig}
\usepackage{float}
\usepackage{algorithm}
\usepackage{algpseudocode}
\floatstyle{plain}
\restylefloat{algorithm}
\usepackage{tabularx}
\usepackage{caption}
\usepackage{booktabs}
\usepackage{listings}
\lstdefinestyle{promptstyle}{
  basicstyle=\scriptsize\ttfamily,
  breaklines=true,
  breakatwhitespace=true,
  columns=fullflexible,
  keepspaces=true,
  showstringspaces=false,
  upquote=true,
  frame=single,
  framesep=2pt,
  xleftmargin=4pt,
  xrightmargin=4pt,
  rulecolor=\color{black!30},
  extendedchars=true,
  inputencoding=utf8,
  literate=%
    {—}{{---}}1
    {–}{{--}}1
    {→}{{->}}2
    {←}{{<-}}2
    {≤}{{<=}}2
    {≥}{{>=}}2
    {±}{{+/-}}3
    {×}{{x}}1
    {°}{{ deg}}4
    {✅}{{[OK]}}4
    {❌}{{[X]}}3
}

\usepackage{array}
\usepackage{booktabs}

\newcolumntype{P}[1]{>{\raggedright\arraybackslash}p{#1}}

\newcommand{\group}[1]{%
  \addlinespace[3pt]
  \specialrule{0.6pt}{0pt}{2pt}
  \multicolumn{3}{@{}l}{\textbf{\textit{#1}}}\\[-1pt]
}

\newcommand{\skill}[1]{\hspace{0.9em}\textit{#1}}

\newcommand{\ours}{SimWorlds}
\newcommand{\bench}{4DBuildBench}

\title{SimWorlds: A Multi-Agent System for Dynamic 3D Scene Creation}

\author{%
  \textbf{Chunjiang Liu$^{1}$ \quad Xiaoyuan Wang$^{1}$ \quad Haoyu Chen$^{2}$ \quad Yizhou Zhao$^{1}$} \\[3pt]
  \textbf{Ming-Hsuan Yang$^{3}$ \quad L\'aszl\'o A. Jeni$^{1}$} \\[6pt]
  {\normalfont\small $^{1}$Carnegie Mellon University \qquad $^{2}$Harvard University \qquad $^{3}$University of California, Merced} \\[5pt]
  {\normalfont\small \url{https://dynsimworlds.github.io}}
}

\begin{document}
\maketitle

\begin{figure}[h]
  \centering
  \includegraphics[width=0.9\linewidth]{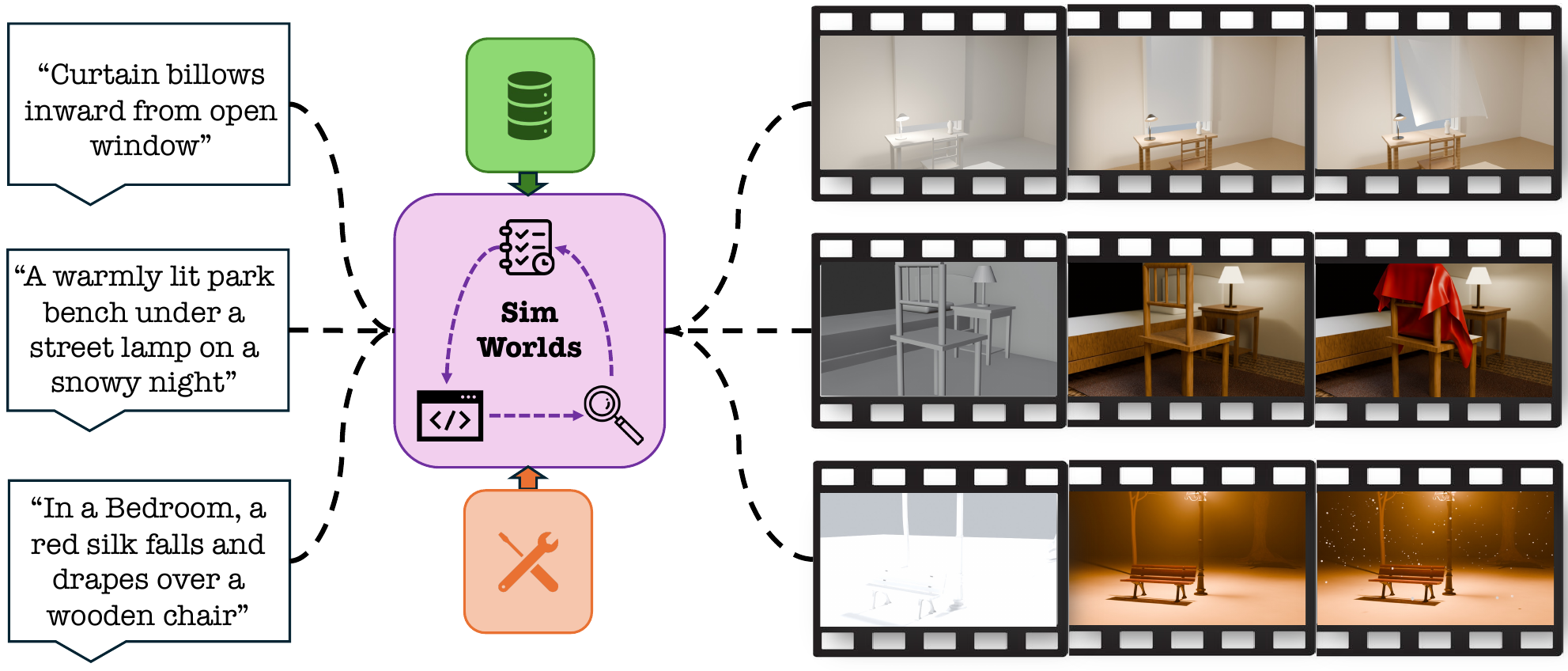}
  \caption{\textbf{\ours{} turns text into dynamic, editable 3D Blender scenes.} Given a natural-language prompt, a planner, coder, and reviewer cooperate to emit a \texttt{.blend} whose geometry, materials, lighting, camera, and motion all remain controllable for downstream editing and reuse.}
  \label{fig:teaser}
\end{figure}

\begin{abstract}
\looseness=-1
LLM agents are increasingly used to translate natural language into 3D scenes in a procedural way, but existing systems focus on static output. Dynamic 4D scenes from text alone, in which liquids flow, particles emit, rigid bodies cascade, and articulated mechanisms move, remain largely unexplored despite their value as editable content and as physics-grounded training data for video generation and embodied AI.
Two challenges set the dynamic case apart from static text-to-scene work: an agent must jointly coordinate spatial layout, multiple physics solvers, temporal sequencing, camera, and lighting in a single coherent scene, and verifying motion correctness from rendered video is fundamentally harder than judging a single image. We present \ours{}: a multi-agent framework that produces dynamic, editable 4D scenes from text, with Blender-specific procedural knowledge, a planner--coder--reviewer workflow driving a fixed ordered sequence of construction stages, a layered scene protocol enforced by a deterministic verifier, and a runtime-state inspection tool suite that catches mechanism failures the rendered image cannot reveal. We also introduce \bench{}, a benchmark for assessing both visual fidelity and physical consistency of the procedural dynamic 3D scenes generated from text prompts. Experiments show that \ours{} outperforms prior dynamic Blender generation baselines.
\end{abstract}

\section{Introduction}
\label{sec:intro}

A modern 3D generative system is increasingly expected to produce more than a visually plausible render.
For downstream graphics, simulation, and content-creation workflows, the desired output is an editable scene artifact: a Blender project in which geometry, materials, lighting, cameras, animation, and physics solvers remain explicit and controllable.
Recent text-to-3D methods have made rapid progress on object-level assets~\citep{poole2023dreamfusion,hong2024lrm,tang2024lgm}, and LLM agents have begun to construct static scenes from natural language~\citep{scenecraft2024,sun20233dgpt,gu2025ll3m,yang2024holodeck,viga2025}.
Generating a dynamic 3D scene from text alone is substantially harder: a dynamic scene must not only look correct in rendered frames but be produced through the correct underlying mechanisms, drawing on rigid-body simulation, cloth, fluids, particles, force fields, deformers, and keyframed control, often combined within a single shot.

This distinction exposes a failure mode largely absent in static generation. In a static scene, visual inspection is a reasonable proxy for correctness; in a dynamic scene, the same rendered video can correspond to very different underlying states: a tablecloth that drapes via a cloth solver, via hand-authored shape keys, or via keyframed mesh edits renders identically, yet only the solver version stays editable, composable, and physically meaningful as the scene changes. Text-to-dynamic-scene generation therefore demands mechanism correctness, not just visual plausibility.

Existing LLM-agent pipelines for Blender are not designed around this requirement. Most render the scene, ask a vision-language model to critique the image, and revise the code, which catches missing objects and obvious material errors but cannot tell whether a fluid domain, properly configured flow and effector objects, and a baked cache exist, or whether the geometry is merely animated. As objects, interactions, and temporal phases multiply, these unchecked failures compound, and the final scene approximates the prompt visually while remaining unusable as a 4D asset.

We formulate dynamic 3D scene generation as plan-grounded, mechanism-aware program synthesis. A text prompt is first converted into an explicit scene plan specifying objects, spatial composition, physical roles and motion phases.
Generation then proceeds through an ordered sequence of typed subtasks, each with its own context, acceptance criteria, and verification.
Crucially, review is not limited to rendered images: the agent inspects the live Blender state, checking whether the expected modifiers are attached, whether physics caches are baked, whether simulated actors move over the intended temporal phase, and whether collision and effector relationships are present.

We instantiate this formulation as \ours{}, a multi-agent framework for text-to-4D scene generation in Blender. A planner compiles the prompt into a structured scene plan; a coder then builds the scene through a fixed ordered sequence of typed stages, each closed by a deterministic verifier that checks the assembled state against a layered scene protocol and a reviewer that judges per-stage criteria, with failed checks triggering localised retries that keep early errors from contaminating later behaviour. Engine-level tools let both the coder and the reviewer read Blender's runtime state: modifier stacks, physics caches, animation channels, and multi-angle previews. A knowledge base auto-derived from upstream Blender sources supplies procedural detail on demand. The result is an editable .blend file whose geometry, materials, lighting, camera, and dynamics remain available for downstream editing, resimulation, and reuse.

We evaluate \ours{} on text-only dynamic scene generation and on multi-step Blender editing via BlenderBench~\citep{viga2025}.
Compared with visual-only agent baselines, \ours{} improves both scene-level correctness and physical integrity, with the gap widening sharply on complex inputs: generation prompts that require multiple interacting objects, long temporal structure, or nontrivial solver configuration, and edit instructions that span several objects or modify physics simultaneously.

Our contributions are:
\begin{itemize}


    \item We present \ours{}, an LLM-agent system that turns text into an editable 4D Blender project: geometry, materials, lighting, cameras, animation, and physics, all controllable.

    \item We build a controllable, physics-grounded generation pipeline that combines a scene protocol and deterministic verifier, render-based review, and an engine-level tool suite.

    \item We introduce \bench{}, 50 scenes across five solver categories (cloth, fluid, rigid body, particle, soft body) and three difficulty levels plus a static category, paired with a two-track evaluation: a deterministic engine-state audit for mechanism correctness, and an itemized VLM judge for whether the prompt's content is visually delivered.
\end{itemize}

\section{Related Work}
\label{sec:related}

\paragraph{Code-Driven and Procedural 3D Scene Generation.}
A growing line of work treats Blender as the runtime for an LLM agent that synthesises scene-construction code. SceneCraft~\citep{scenecraft2024} and 3D-GPT~\citep{sun20233dgpt} translate text into Blender scripts coordinated through relational scene graphs. BlenderAlchemy~\citep{sun2024blenderalchemy} iteratively refines materials under VLM feedback. LL3M~\citep{gu2025ll3m} composes planner, retrieval, and coder agents over a BlenderRAG knowledge base, and reports object-level results at high quality. A complementary thread bypasses learned generation entirely: procedural pipelines such as Infinigen~\citep{raistrick2023infinigen} hand-craft generators for Blender-rendered nature scenes, and ProcTHOR~\citep{deitke2022procthor,ge2026airsim360,lin2026depthany} programmatically synthesises indoor environments for embodied agents. Layout-generation methods predict object placements from text, scene graphs, or partial context~\citep{paschalidou2021atiss,fang2023ctrlroom,feng2024layoutgpt,lin2024instructscene,ocal2024sceneteller,tang2024diffuscene}, typically retrieving furniture from large asset libraries~\citep{deitke2023objaverse,deitke2023objaversexl}. These systems share our artifact target, an editable, code-defined Blender file, but their headline results target static objects or single-room layouts; the dynamic regime, in which motion and physics are first-class outputs, has not been demonstrated end-to-end from text.

\paragraph{Render-Inspect Loops and Editing Benchmarks.}
A parallel thread closes the verification loop by rendering the in-progress scene and asking a VLM to identify discrepancies. VIGA~\citep{viga2025} formalises this as a code-render-inspect loop and is the closest published system to ours; it conditions on a reference image of the target and reports a 4D mode through qualitative figures only. Alongside its system, VIGA introduces BlenderBench, an open-ended editing suite of 27 tasks spanning spatial adjustments, progressive editing, and compositional generation, on which existing one-shot baselines remain far below human performance. A separate effort, BlenderGym~\citep{gu2025blendergym}, contributes 245 handcrafted editing tasks and explicitly identifies a class of failures uncaught by its own photometric and CLIP-based metrics: scenes that match the goal pixels through the wrong mechanism. \ours{} retains the iterative-verification design but drives it with both rendered previews and mechanism-level signals (bake state, modifier stack, fcurves, motion deltas) read from the engine itself, and extends the loop to text-only 4D. Our edit mode runs unchanged on BlenderBench, enabling a direct comparison with VIGA on the benchmark it introduced.

\paragraph{Neural Text-to-3D and Text-to-4D.}
\looseness=-1 A separate paradigm produces 3D and 4D content as neural representations rather than as graphics-engine code. Object-level methods distil 2D priors into NeRFs or 3D Gaussians~\citep{poole2023dreamfusion,lin2023magic3d,chen2023fantasia3d,wang2024prolificdreamer,hong2024lrm,tang2024lgm,xu2024instantmesh,kerbl2023gaussian,tang2024dreamgaussian,yi2024gaussiandreamer,chen2024gsgen,gao2024graphdreamer}, with multi-view and image-conditioned variants resolving Janus-like inconsistencies~\citep{liu2023zero123,shi2024mvdream,long2024wonder3d}. Scene-level optimisation extends this to layouts, rooms, and gallery environments~\citep{cohen2023setthescene,zhang2024scenewiz3d,hoellein2023text2room,fridman2023scenescape,zhou2024gala3d,li2024director3d}. 4D variants animate the representation with video priors~\citep{singer2024mav3d,bahmani20244dfy,ren2024dreamgaussian4d,xu2024comp4d,ren2024l4gm,bahmani2024tc4d,liang2024diffusion4d,xie2024sv4d,wang2025holigs}, themselves drawing on text-to-image and text-to-video diffusion~\citep{rombach2022high,ho2022video,blattmann2023svd,singer2023makeavideo,brooks2024sora,liu2026omniroam,wang2026panoworld}. Physics-aware methods attach material parameters to existing fields~\citep{physdreamer2024,xie2024physgaussian}, while system-identification methods recover per-object physical parameters from video through differentiable simulation or learned neural constitutive models~\citep{liu2026mosiv,zhao2025masiv}. This output is visually compelling but is not our target artifact: it cannot be opened in Blender, resimulated, or composed with downstream tools, and its motion is learned rather than solved. Kubric~\citep{greff2022kubric} renders programmatic Blender physics but is configured by code, not natural language; none of these produce the editable, physics-driven scenes in Blender that our system targets.

\paragraph{LLM Agents and Long-Horizon Execution.}
\looseness=-1 Outside graphics, a parallel line of work studies what makes LLM agents reliable on long-horizon tasks. ReAct~\citep{yao2023react} and Reflexion~\citep{shinn2023reflexion} alternate reasoning with environment feedback; Self-Refine~\citep{madaan2023selfrefine} formalises iterative revision, while \citet{huang2024selfcorrect} caution that LLMs cannot reliably self-correct without external grounding; Voyager~\citep{wang2023voyager}, Toolformer~\citep{schick2023toolformer}, CRITIC~\citep{gou2024critic}, and CodeAct~\citep{wang2024codeact} ground critique and action in tool calls and code; AutoGen~\citep{wu2023autogen}, MetaGPT~\citep{hong2024metagpt}, and ChatDev~\citep{qian2024chatdev} factorise tasks across role-specialised agents. Recent perspectives crystallise the load-bearing components as context engineering~\citep{anthropic2025context,mei2025contextsurvey}, tool design~\citep{anthropic2025tools}, and plan-grounded execution~\citep{weng2023agents,park2023generativeagents}. \ours{} adopts these as its spine; the contribution is not the principles but their instantiation for 4D scene generation in Blender: context scoped to typed stages, tools that read runtime state alongside previews, and a plan whose physics commitments are reconciled against the engine after every step.


\section{Method}
\label{sec:method}

\begin{figure}[t]
  \centering
  \includegraphics[width=\linewidth]{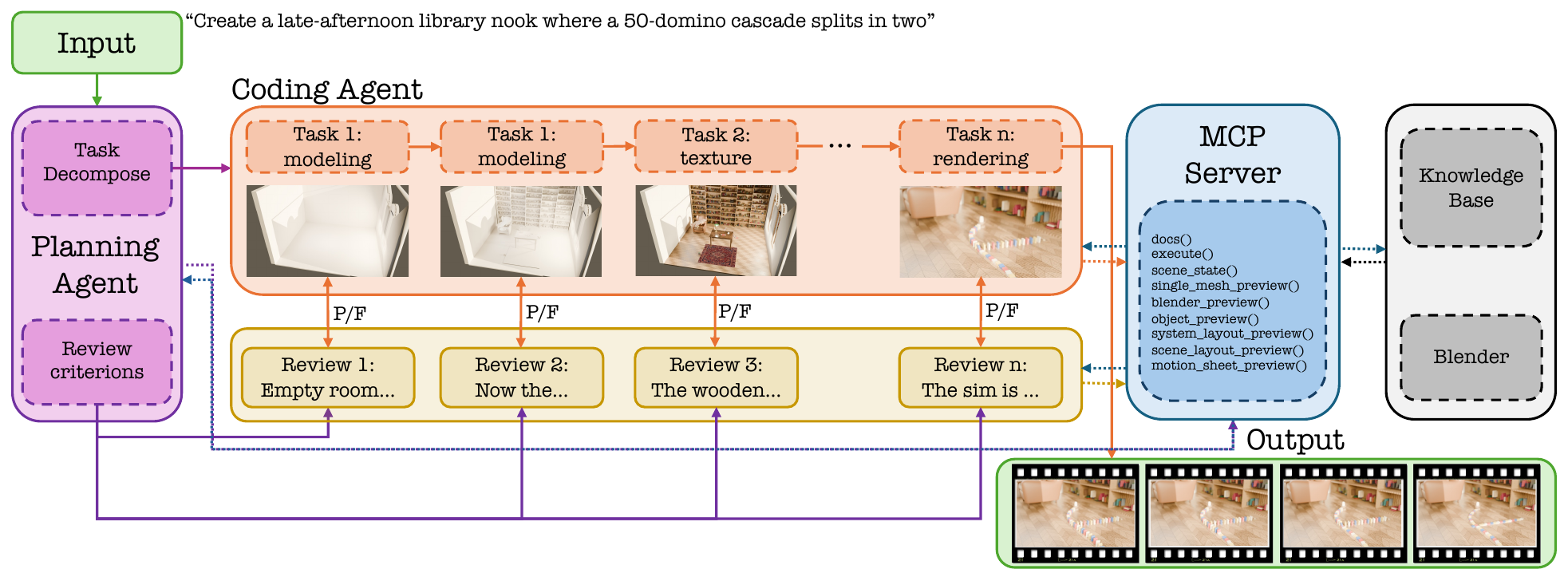}
  \caption{\textbf{Pipeline overview.} The planner compiles the prompt into a single scene plan; construction then proceeds through a fixed stage sequence, each stage running the coder, a deterministic verifier (scene protocol, \S\ref{sec:protocol}), and a reviewer, with failed checks triggering bounded local retries. Once all stages close, the scene is rendered as a frame-sequence video.}
  \label{fig:method}
\end{figure}

\ours{} is organised around one idea: a correct render does not guarantee a correctly built scene, so every construction step is validated against Blender's engine state rather than its rendered image, and the scene is assembled through a fixed, checkable sequence of stages so that structural and mechanism errors are caught deterministically and early. Concretely, \ours{} turns a text prompt into a dynamic, editable scene by separating one-shot planning from a stage-by-stage execute--verify--review loop (Alg.~\ref{alg:pipeline}).
The planner first converts the prompt into a global scene plan that specifies objects, their spatial layout as a typed relation graph, physical properties, motion phases, and rendering intent.
Generation then proceeds through a fixed ordered sequence of construction stages (modeling, UV, texture, deformation setup, motion, camera, light, render); for each, the planner emits a tactical plan against the scene plan, and may opt a stage out for scenes that do not need it.
At each stage the coder writes a bpy script that extends a running scene, an orchestrator-side verifier mechanically checks the resulting Blender state against a fixed protocol, and a reviewer agent then judges per-stage acceptance criteria from rendered previews and runtime readouts before the pipeline advances.
Figure~\ref{fig:method} sketches the full loop, which rests on two core mechanisms (each validated by its own ablation, \S\ref{sec:ablation}) and two enabling components.
The first core mechanism is a staged construction pipeline (\S\ref{sec:pipeline}) that anchors every coder turn to a single structured specification rather than to an open-ended task list, so retries and reviews are well-scoped.
The second is a scene protocol layered on Blender's natives and the orchestrator-side verifier that enforces it (\S\ref{sec:protocol}); together they turn assembly correctness into a deterministic check, catching structural failures (e.g.\ missing parent chains, ungrounded objects, undeclared interpenetrations) before any visual verification runs.
Two enabling components complete the loop: a tool suite (\S\ref{sec:tools}) that gives the verifier and reviewer direct runtime-state readouts and multi-angle previews, and an auto-derived knowledge base (\S\ref{sec:knowledge}) that supplies Blender-specific procedural knowledge per stage.

\subsection{Staged Construction Pipeline}
\label{sec:pipeline}
\ours{} builds on a fixed pipeline of construction stages $\mathcal{S}$: modeling, UV, texture, deformation setup, motion, camera, light, and render, applied in the same order to every scene. The order follows the dependency structure of a Blender scene: materials depend on UVs, deformers on geometry, motion on rigging, and lighting and camera on the assembled scene. The planner emits a single global scene plan once, then a per-stage tactical plan at each stage entry, and may opt a stage out when a scene does not need it (e.g.\ UV/texture for primitive geometry, deformation setup for fully rigid scenes). Staging this way keeps retries, checkpoints, and reviewer scope well-defined per stage, while the global scene plan stays immutable across the run (Alg.~\ref{alg:pipeline}).

\begin{algorithm}[t]
\small
\hrule\vspace{2pt}
\noindent\textbf{\ours{} staged construction loop.}\vspace{3pt}\hrule\vspace{3pt}
\algtext*{EndIf}\algtext*{EndFor}
\begin{algorithmic}[1]
\Require prompt $q$; fixed stage sequence $\mathcal{S} = (s^1, \ldots, s^N)$; initial scene state $\sigma_0 = \emptyset$
\State $\mathcal{P} \leftarrow \textsc{StrategicPlanner}(q)$ \Comment{scene plan: scene specification + relation graph + motion phases + per-stage opt-outs}
\For{$k = 1$ to $N{-}1$} \Comment{construction stages (render handled separately)}
    \If{$s^k \in \mathcal{P}.\texttt{opt\_out}$} $\sigma_k \leftarrow \sigma_{k-1}$; \textbf{continue} \Comment{skipped stage carries state forward} \EndIf
    \State $f \leftarrow \emptyset$ \Comment{feedback: verifier + reviewer reports}
    \Repeat
        \State $\tau_k \leftarrow \textsc{TacticalPlanner}(s^k, \mathcal{P}, \sigma_{k-1})$
        \State $\sigma_k \leftarrow \textsc{Exec}(\textsc{Coder}(\tau_k, \sigma_{k-1}, f))$
        \State $v_k \leftarrow \textsc{Verifier}(\sigma_k, \mathcal{P})$ \Comment{deterministic: scene protocol (\S\ref{sec:protocol})}
        \State $\rho_k \leftarrow \textsc{Reviewer}(\tau_k, \mathcal{P}, \sigma_k)$ \Comment{perceptual: per-stage criteria}
        \State $f \leftarrow (v_k, \rho_k)$;\ \ $d \leftarrow \textsc{Planner}(v_k, \rho_k)$ \Comment{advance / retry / replan}
    \Until{$d = \texttt{advance}$ \textbf{ or } retries/replans exhausted ($\Rightarrow$ \textbf{abort})} \Comment{retries/replans bounded; see App~\ref{app:implementation}}
    \State $\textsc{Checkpoint}(\sigma_k, s^k)$
\EndFor
\State $\mathcal{O} \leftarrow \textsc{Render}(\sigma_{N-1})$ \Comment{frame sequence}
\State $\rho^{\star} \leftarrow \textsc{FinalReviewer}(\mathcal{P}, \mathcal{O})$
\If{$\rho^{\star} = \texttt{needs\_fix}$}
    \State $\sigma_{N-1} \leftarrow \textsc{Exec}(\textsc{Coder}(\rho^{\star}, \sigma_{N-1}))$;\ \ $\mathcal{O} \leftarrow \textsc{Render}(\sigma_{N-1})$
\EndIf
\State \Return $(\mathcal{O}, \sigma_{N-1})$
\end{algorithmic}
\vspace{2pt}\hrule
\caption{The strategic planner emits one global scene plan; construction then runs the fixed stage sequence. Each stage repeats coder, deterministic verifier (scene protocol, \S\ref{sec:protocol}), and reviewer until the planner advances it, so a hard verifier or reviewer failure cannot pass; the stage aborts once its retry and replan budgets are exhausted. After all stages close, the scene is rendered and a final reviewer judges the output.}
\label{alg:pipeline}
\end{algorithm}

\paragraph{Scene Plan.}
The scene plan consists of three parts: a scene specification (objects with dimensions, PBR material, position, and physics role; a typed relation graph over objects encoding spatial assembly; coordinated groups; lighting; camera; render), a motion plan for dynamic scenes, and a list of stages to opt out for the current scene. Full field schemas are listed in Appendix~\ref{app:implementation}.

\paragraph{Per-Stage Execution Loop.}
At stage $s^k$, the orchestrator opens a fresh coder and reviewer session, while the planner persists across the whole run, and drives a per-stage state machine. The tactical planner first emits a plan $\tau_k$ specifying the assets to realise, the relations to satisfy, the per-stage acceptance criteria, and the perceptual judgments the reviewer should answer. The coder extends the live scene with a bpy script; the verifier of \S\ref{sec:protocol} checks the assembled state; and the reviewer judges $\tau_k$'s perceptual criteria from rendered previews and the structured scene readout, additionally consuming per-phase motion evidence on motion stages. On this evidence the planner chooses one of four transitions: it can advance to the next stage; retry, having the coder reuse its session to fix the current attempt from the verifier and reviewer feedback; replan, revising the tactical plan itself; or abort. A hard verifier failure or a blocking review can never advance: the orchestrator forces a retry, so the stage loops through coder, verifier, reviewer, and decision until it passes, and retries and replans stay bounded.

\paragraph{Final Render and Review.}
Once every construction stage closes, the runner renders the scene as a single frame for static scenes or a frame-sequence video for dynamic ones, and a separate final reviewer judges the rendered result against the plan. A failed verdict rolls back to the last stage checkpoint, re-invokes the coder once with consolidated fix instructions, and re-renders.

\paragraph{Edit Mode.}
The same loop accommodates edit-style prompts at no architectural cost.
Given an existing .blend file, a natural-language instruction, and an optional reference image of the target, the runner loads the file, populates $s_0$ from the file's scene state, attaches the reference image as additional planner context, and asks the planner to decompose the request into edit-only tasks (additions, modifications, removals) rather than redoing static elements that already exist correctly.

\subsection{Scene Protocol and Verifier}
\label{sec:protocol}

Blender's native scene graph (parent--child, collections, modifiers, drivers, constraints) expresses transform inheritance, organisation, and runtime dependencies, but it does not express assembly correctness. There is no native primitive for ``these meshes form one logical object,'' ``surface $A$ touches surface $B$,'' or ``this object is grounded.'' Without these notions every coder turn must be re-validated by visual inspection, and the failure modes the rendered image cannot show (a chair leg floating 5\,mm above the floor, a tabletop interpenetrating a wall, an object never parented to anything) silently accumulate. \ours{} layers a small, declarative protocol on top of Blender's natives and enforces it with an orchestrator-side verifier that runs automatically after every coder turn, catching structural and mechanism errors before they ever reach a render.

\paragraph{Protocol structure.}
The protocol imposes a strict three-level containment hierarchy, built entirely from Blender's native collections, custom properties, and parenting. L1 is the scene as a whole. Each L2 inside it is a system grouping: a named set of related objects, such as one room, or a holder of scene-level state, such as the lights and cameras. Each L3 inside an L2 is a single logical object, however many meshes it is built from: it gathers all of those meshes under one root Empty, a geometry-free anchor object, and parents every mesh to it, so the object moves and is checked as a unit. Since a Blender collection carries no inherent role, a custom property tags each one as an L2 or an L3; collections left untagged are ordinary organisational collections that lie outside the protocol.

On top of this hierarchy the protocol makes explicit two object-to-object relations that Blender's scene graph leaves implicit. The first is surface contact, where one object rests on or against another, such as a chair leg on the floor; it is declared on either of the two objects and the verifier later confirms that the surfaces actually meet. The second is co-movement, where one object must travel with another, such as a sword carried in a hand; it is expressed through Blender's native parent--child constraints.

\paragraph{Bipartite plan graph.}
\looseness=-1 The planner commits the scene's intended assembly up front as a bipartite plan graph: one set of nodes for the objects, another for the relations among them. Each relation node carries a typed spatial relation and the objects it relates, spanning support, containment, orientation, and regular arrangement, such as one object resting on another or several arranged in a ring (the full vocabulary and its rules are in Appendix~\ref{app:protocol}). Because relations are nodes, not edges, one relation can span several objects at once. The coder realises this graph as the collection tree and the contact declarations above, giving the verifier a stated assembly intent to check.

\paragraph{Rule families.}
\looseness=-1 The verifier runs four families of deterministic rules against the live Blender state. Structural rules C1--C6 check the collection layout, and G1--G3 check the anchor-and-parent chains within each object. Geometric rules V1--V2 are BVH distance tests confirming that every declared contact pair actually touches and that no undeclared pair interpenetrates, which catches floating contacts and accidental punch-through. Soft rules W1--W5 raise warnings about grounding, within-object connectivity, and missing cameras or lights. Plan-vs-state rules R1--R18, with seven further motion-timing rules, re-check the planner's relation graph and motion phases against the realised geometry, for example that a declared circular leg arrangement is realised to tolerance and that a settle phase comes to rest in its final frames. The full catalogue and BVH-test details are in Appendix~\ref{app:protocol}.

\subsection{Tool Suite}
\label{sec:tools}

Verification in prior Blender-agent systems~\citep{scenecraft2024,sun20233dgpt,gu2025ll3m,viga2025} reduces to visual critique: render, feed to a VLM, request issues.
This is often enough for static scenes but cannot tell whether a dynamic effect uses the right mechanism or is merely faked to look identical.
\ours{} replaces visual self-critique with direct inspection of Blender's runtime state, through a tool suite (full inventory in Appendix Table~\ref{tab:tools}) grouped into state observation, modification, and knowledge access. Three tools supply the evidence frames cannot: blender\_scene\_state returns a structured readout of the live scene (collections, modifier stacks, physics caches, fcurve channels), showing what was actually built rather than rendered; a multi-granularity preview family renders a mesh, an L3 object, an L2 system, or the whole scene from fixed multi-angle presets; and blender\_motion\_sheet\_preview samples a per-actor frame strip annotated with phase boundaries and bake state.

\subsection{Knowledge}
\label{sec:knowledge}

\looseness=-1 4D Blender content draws on too many subsystems (meshes, PBR shaders, the rigid-body world, cloth and fluid solvers, particle systems, keyframe animation, the compositor) for any single system prompt to cover. \ours{} therefore exposes knowledge through a single tool, blender\_docs(query), backed by a knowledge base auto-derived per Blender version from upstream sources: the bl\_rna schemas of every bpy.types class, every bpy.ops operator's signature, the official Python API guides, and the full Blender manual. A query resolves to an exact schema, a manual page, or a ranked candidate list; the base regenerates automatically when the Blender version changes, with no LLM in the build except a content-addressed page-summary step.


\section{Experiments}
\label{sec:experiments}
\label{sec:benchmark}

\subsection{Setup}

\paragraph{Benchmark.}
We evaluate \ours{} on \bench{}, a 50-scene benchmark of self-contained prompts for text-to-4D Blender scene generation. \bench{} is organised along two axes: the \emph{mechanism} an artifact must use, and the \emph{difficulty} of authoring it correctly. The mechanism axis covers the five core Blender physics solvers (cloth, fluid, rigid body, particle systems, and soft body), each as a category of 9 prompts, plus a static category of 5 furnished-scene prompts that exercise object inventory and spatial layout. The difficulty axis defines three levels per dynamic category, three prompts each, escalating from a single solver actor (D1) to within-category complexity (D2, e.g.\ self-collision or force fields) to cross-category interaction in one shot (D3, e.g.\ a rigid block crushing a soft-body slab); per-level definitions and example prompts are in Appendix~\ref{app:benchmark}. Editing is evaluated separately on the BlenderBench~\citep{viga2025} via our edit mode (\S\ref{sec:pipeline}).

\paragraph{Evaluation Metrics.}
The central failure mode in dynamic-scene generation is a scene that looks right but is built through the wrong mechanism. A purely visual judge cannot see this while an engine-state check cannot see whether the prompt's objects and motions are visually delivered. \bench{} therefore scores each scene on two complementary tracks, detailed in Appendix~\ref{app:rubric}.

\emph{(1) Engine-state audit.} A deterministic audit runs in headless Blender and reads the scene's runtime state directly: solver modifiers, baked caches, collision partners, constraints, force fields, and the keyframe density that betrays faked motion. It checks this state against a per-scene ground-truth mechanism specification drawn from a library of 42 typed predicates (e.g.\ \emph{cloth modifier present and cache baked}; \emph{static scene carries no baked physics caches}). We report two aggregates: \textbf{MPR} (Mechanism Pass Rate), the per-scene mean over actors of the fraction of each actor's predicates satisfied; and \textbf{SPR} (Structural Pass Rate), the per-scene fraction of declared spatial relations that hold geometrically under a BVH surface-distance test. The audit shares no code with \ours{}'s in-loop verifier and reads the .blend alone; where a baseline emits no logical-object grouping, SPR infers each object's assembly geometrically, scoring all systems on identical terms.

\emph{(2) Itemized VLM judge.} Building on VLM- and LLM-as-judge evaluation for generative vision~\citep{zheng2023judging,hu2023tifa,he2024videoscore,chen2026memobench}, a GPT-5.5 judge receives the prompt and five frames sampled uniformly across the clip ($t \in \{0,25,50,75,100\}\%$) and scores five dimensions: objects present, spatial relations, actions visible, visual quality, and aesthetics. Rather than emit numeric scores, it enumerates atomic items from the prompt (one per named object, relation, and action) and returns a binary verdict per item; the dimension score is the fraction that hold, giving partial credit and a concrete evidence trail. Mechanism realism is left to the audit track, since vision models judge it unreliably~\citep{bansal2024videophy}.

\paragraph{Baseline.}
We compare against VIGA~\citep{viga2025}, the only prior published system with a dedicated dynamic-scene mode for from-scratch 4D scene generation in Blender. VIGA exposes a dynamic-scene mode alongside its static-scene mode, runs a dual-agent generator--verifier loop, and emits an editable Blender file, which makes it directly runnable on \bench{} prompts. It also shares the dual-agent shape with \ours{}, so the comparison isolates the contributions of the staged construction pipeline, the scene protocol and verifier, and the engine-level tool suite.\footnote{Other recent LLM-driven 3D scene systems target adjacent settings and are not directly runnable on \bench{}. LL3M~\citep{gu2025ll3m} routes generation through a closed cloud server with no code or model release. SceneCraft~\citep{scenecraft2024} produces static layouts only, with no physics or temporal output. Holodeck~\citep{yang2024holodeck} outputs AI2-THOR rooms, a different artifact type. BlenderGym~\citep{gu2025blendergym} methods operate in editing mode against reference images. Concurrent VoxelCodeBench~\citep{voxelcodebench2026} targets static voxel construction in Unreal.}

\subsection{Results}

\paragraph{Quantitative Results.}
We evaluate \ours{} and VIGA on \bench{}, and score both with the engine-state audit and the itemized VLM judge. The results are reported in Table~\ref{tab:main}.
\begin{table}[H]
\centering
\small
\setlength{\tabcolsep}{5.5pt}
\begin{tabular}{@{}l cc cc cc@{}}
\toprule
 & \multicolumn{2}{c}{\textbf{MPR}$\uparrow$} & \multicolumn{2}{c}{\textbf{SPR}$\uparrow$} & \multicolumn{2}{c}{\textbf{VLM score}$\uparrow$} \\
\cmidrule(lr){2-3}\cmidrule(lr){4-5}\cmidrule(lr){6-7}
\textbf{Category} & \ours{} & VIGA & \ours{} & VIGA & \ours{} & VIGA \\
\midrule
Cloth      & \textbf{0.92} & 0.85 & \textbf{0.78} & 0.72 & \textbf{0.82} & 0.81 \\
Fluid      & \textbf{0.95} & 0.56 & \textbf{1.00} & 0.89 & 0.75 & 0.75 \\
Rigid body & \textbf{0.83} & 0.61 & \textbf{0.83} & 0.67 & 0.80 & \textbf{0.82} \\
Particle   & \textbf{0.79} & 0.62 & \textbf{0.89} & 0.67 & \textbf{0.84} & 0.70 \\
Soft body  & \textbf{0.87} & 0.72 & \textbf{0.89} & 0.72 & \textbf{0.89} & 0.87 \\
Static     & -- & -- & \textbf{0.98} & 0.42 & \textbf{0.79} & 0.67 \\
\midrule
\textbf{Overall} & \textbf{0.87} & 0.67 & \textbf{0.89} & 0.70 & \textbf{0.82} & 0.78 \\
\bottomrule
\end{tabular}
\caption{Per-category results on \bench{}. \textbf{Overall} is the macro average over cells: 15 dynamic cells for MPR and all 17 for SPR and VLM score.}
\label{tab:main}
\end{table}
The results show that MPR is where \ours{} and VIGA diverge most decisively ($0.87$ vs $0.67$), while VLM stays comparable ($0.82$ vs $0.78$). The reason is that the VLM judge scores only a few frames sampled at fixed intervals, which cannot reveal whether the motion across them is correct, so a keyframed or shape-key scene can be wrong yet score well when each sampled frame looks right on its own; the engine-state audit instead reads the solver state behind those frames and rejects it. The same blind spot is what lets VIGA's visual-only verifier accept such fakes during generation. SPR moves in the same direction ($+0.19$), but because it scores spatial structure against object groupings that \ours{} emits and baselines do not, its cross-system fairness is limited; we mitigate the confound geometrically (Appendix~\ref{app:rubric}) and read SPR as supporting evidence.

\begin{figure}[H]
  \centering
  \includegraphics[width=\linewidth]{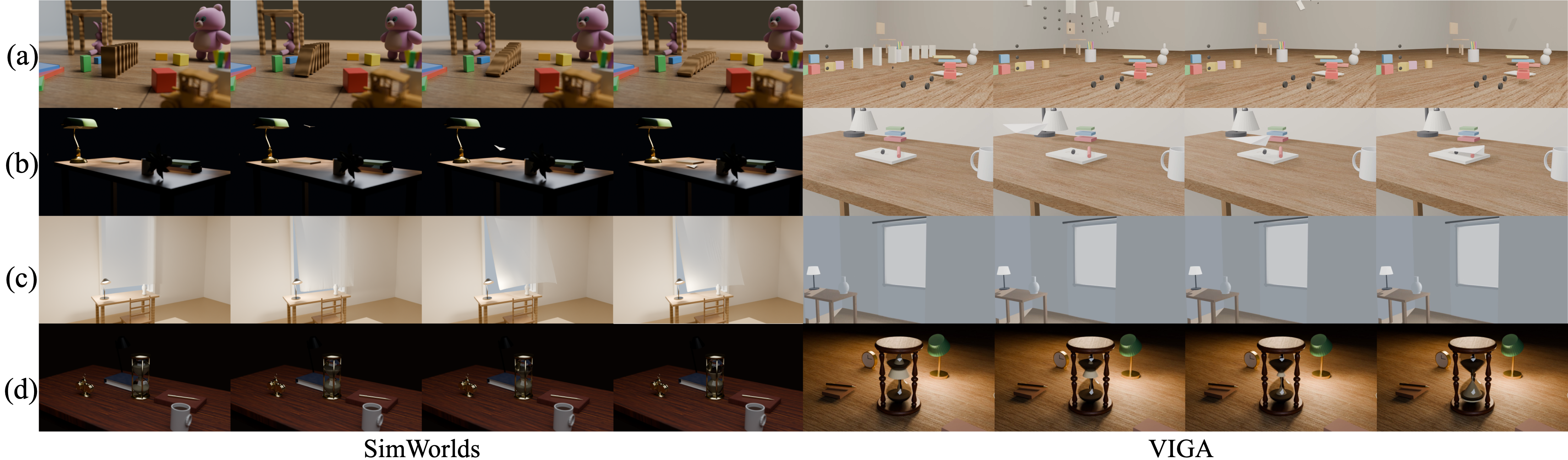}
  \caption{Qualitative comparison across four dynamic sequences: (a) a domino cascade in a child's bedroom; (b) a paper airplane landing on a cluttered desk; (c) a wind-blown curtain billowing over a desk; (d) a glass hourglass pouring sand from its upper to its lower bulb.}
  \label{fig:gallery}
\end{figure}

\paragraph{Qualitative Results.}
As shown in Fig~\ref{fig:gallery}, across all four sequences \ours{} realises the correct mechanism while VIGA fails it. The common cause is architectural: VIGA closes its loop with a VLM verifier that inspects sampled frames, which is insufficient to tell whether motion or physics is actually correct, so a scene whose sampled frames look right is accepted however its dynamics were produced. \ours{} instead verifies each stage against engine state through its protocol verifier (\S\ref{sec:protocol}), so that a faked keyframe is caught at construction.

\paragraph{Scene Editing.} \ours{} runs in edit mode (\S\ref{sec:pipeline}) on VIGA's BlenderBench~\citep{viga2025} without architectural change, at a matched 6-iteration budget on Opus~4.7 and scored by VIGA's own PL, N-CLIP, and VLM metrics (Table~\ref{tab:blenderbench}; full setup in Appendix~\ref{app:blenderbench}). Both systems receive the same input, an existing scene with a text instruction and a target image. VIGA edits through a single generator-verifier loop that re-examines the whole scene each iteration; \ours{} instead decomposes the request and routes it through the staged pipeline. This precise localisation, together with the planning-coding-reviewing loop carried over from the generation pipeline (\S\ref{sec:pipeline}, \S\ref{sec:tools}), leads to \ours{}'s improved performance over VIGA across all difficulty levels on BlenderBench.
\begin{table}[H]
\centering
\small
\begin{tabularx}{\linewidth}{@{}l*{6}{>{\centering\arraybackslash}X}@{}}
\toprule
& \multicolumn{3}{c}{\textbf{VIGA~\citep{viga2025}}} & \multicolumn{3}{c}{\textbf{\ours{}}} \\
\cmidrule(lr){2-4}\cmidrule(lr){5-7}
& PL$\downarrow$ & N-CLIP$\downarrow$ & VLM$\uparrow$ & PL$\downarrow$ & N-CLIP$\downarrow$ & VLM$\uparrow$ \\
\midrule
Level 1 & 7.60 & 6.06 & 2.97 & \textbf{6.53} & \textbf{5.91} & \textbf{3.56} \\
Level 2 & 2.44 & 3.27 & 2.97 & \textbf{0.97} & \textbf{3.07} & \textbf{3.84} \\
Level 3 & 7.63 & 6.27 & 2.36 & \textbf{5.99} & \textbf{5.23} & \textbf{2.89} \\
\midrule
Overall & 6.02 & 5.28 & 2.76 & \textbf{4.63} & \textbf{4.80} & \textbf{3.41} \\
\bottomrule
\end{tabularx}
\caption{Scene-editing results on BlenderBench.}
\label{tab:blenderbench}
\end{table}

\subsection{Ablations}
\label{sec:ablation}

We ablate the two mechanisms \ours{} adds to a plain planner--coder--reviewer loop, the scene protocol and deterministic verifier (\S\ref{sec:protocol}), and the staged construction pipeline (\S\ref{sec:pipeline}). Experiments are conducted on a 15-scene subset, whose absolute scores sit above the full-50 numbers of Table~\ref{tab:main}.

\begin{table}[H]
\centering
\small
\begin{tabular}{@{}lccc@{}}
\toprule
\textbf{Configuration} & \textbf{MPR}$\uparrow$ & \textbf{SPR}$\uparrow$ & \textbf{VLM}$\uparrow$ \\
\midrule
\ours{} (full)          & \textbf{0.97} & \textbf{0.97} & \textbf{0.87} \\
\quad Ab1: no verifier  & 0.93 & 0.87 & 0.84 \\
\quad Ab2: no stages    & 0.94 & 0.92 & 0.76 \\
\bottomrule
\end{tabular}
\caption{Ablations results on a fixed 15-scene subset of \bench{}: one scene per mechanism category $\times$ difficulty level.}
\label{tab:ablation}
\end{table}

On this 15-scene subset the two ablations show a double dissociation. Removing the verifier hurts SPR most: without the deterministic gate, structural breaks reach the final scene uncorrected instead of triggering a retry, most visibly on the multi-object D3 scenes, two of which drop to zero SPR. Removing staged construction hurts VLM most: a single pass still assembles the right objects but loses the per-stage review that keeps materials, lighting, and composition on track. MPR is robust to both, since the solver setup itself is authored reliably regardless. Each mechanism thus guards a different axis, structural correctness and visual polish, and neither is redundant with the other.


\section{Limitations}
\label{sec:limitations}

\ours{} grounds mechanism and geometry against the engine, but its perceptual judgments, such as whether the plan follows the prompt, whether objects are arranged sensibly, and whether the scene composes well, still rest on the LLM and VLM rather than the deterministic checks, and are correspondingly less reliable. \ours{} is also text-only; conditioning on a reference image to supervise layout and appearance is a promising direction for future work.

\section{Conclusion}
\label{sec:conclusion}

We presented \ours{}, a multi-agent framework for text-to-4D scene generation in Blender that emits editable .blend files whose dynamics are realised by physics solvers rather than imitated by hand-authored animation, together with \bench{}, a benchmark that scores generated scenes on both visual quality and mechanism correctness. \ours{} structures generation as a fixed staged pipeline that builds and verifies the scene one stage at a time, so errors are caught and repaired locally instead of accumulating across the build. A lightweight scene protocol keeps the generated assets well-organised and verifiable, leaving the VLM reviewer the perceptual judgments, such as aesthetic composition and prompt alignment, that deterministic checks cannot make. We see \ours{} as a step toward procedural agents whose output is not a piece of media but an editable, physics-driven asset that others can pick up, perturb, and build on.


\section*{Acknowledgements}
    YZ was supported in part by the SoftBank Group–ARM Fellowship.

\newpage
\bibliographystyle{unsrtnat}
\bibliography{references}

\newpage
\appendix
\section{Implementation Details}
\label{app:implementation}

\paragraph{Stack.}
\ours{} uses \texttt{claude-opus-4-7} as the underlying model for all three agent roles (planner, coder, and reviewer). Evaluation uses two independent scorers (Appendix~\ref{app:rubric}): a deterministic engine-state audit that runs inside headless Blender with no LLM in the loop, and an itemized VLM judge on \texttt{gpt-5.5} run in a separate session; the BlenderBench editing comparison instead uses a \texttt{gpt-4o} judge (Appendix~\ref{app:blenderbench}). Each judge scores every scene once at the provider's default temperature, with no multi-seed averaging, and the judge prompts are released verbatim under \texttt{src/bench/}. The audit's predicate scoring uses no LLM; only an optional actor-matching fallback uses \texttt{claude-sonnet-4-6}. Blender 5.1 is exposed through an MCP server that provides the tool-use APIs of \S\ref{sec:tools}. End-to-end runtime is approximately 70--120 minutes per scene on a single workstation (Apple M4, 16~GB unified memory), dominated by Blender bake time on dynamic prompts; the LLM is consumed via API and contributes negligible local compute.

\paragraph{Scene Plan.}
The planner writes a structured plan in JSON and never operates Blender directly: it emits one \emph{strategic plan} at the start of a run and a per-stage \emph{tactical plan} at each stage entry, then issues an advance/retry/replan/abort decision after each stage. The plan's load-bearing commitments are declared as custom properties on each logical object's L3 collection: a \texttt{deformation\_kind} $\in$ \{\texttt{rig}, \texttt{sim\_cloth}, \texttt{sim\_fluid}, \texttt{sim\_rigid}, \texttt{none}\}, set at the \texttt{deformation\_setup} stage, and a \texttt{motion\_kind} $\in$ \{\texttt{keyframe}, \texttt{bake}, \texttt{none}\}, set at the \texttt{motion} stage. These, together with the spatial-relation graph of Appendix~\ref{app:protocol}, fix what the verifier's R and T rules check. A stage left undeclared is a zero-cost no-op (the R/T rules self-gate on the declaration), so a static scene passes through \texttt{deformation\_setup} and \texttt{motion} cleanly. The planner sets no bpy implementation specifics, which API call to use or which value to tune; that is the coder's job. The full plan schema and planner prompt are in the released repository (\texttt{prompts/}).

\paragraph{Per-Stage Loop Implementation.}
Each stage runs the coder, the deterministic verifier, and the per-stage reviewer until the planner advances it. The coder is allowed up to \texttt{MAX\_RETRIES\_PER\_STAGE}$=10$ coder retries and \texttt{MAX\_REPLANS\_PER\_STAGE}$=5$ tactical replans per stage before the stage aborts. Each \texttt{blender\_execute} call runs in a fresh Python namespace; persistent state lives in the live Blender scene (\texttt{bpy.data}), which later stages build on rather than rebuild. Every closed stage writes a checkpoint \texttt{checkpoints/\textless stage\textgreater.blend}, so a coder failure rolls the scene back to the previous stage's checkpoint.

\paragraph{Agents.}
The pipeline runs four roles: a \emph{planner} (one persistent session across the run, reasoning over plans in JSON with no Blender tools), a \emph{coder} (fresh session per stage, persisting across that stage's retries), a \emph{per-stage reviewer} (fresh session per stage), and a \emph{final reviewer} (one-shot, on the rendered output). Each role prompt is composed at runtime from the role file plus the shared scene-protocol documentation (\texttt{src/agent/prompts/scene\_protocol.md} and \texttt{docs/scene\_protocol.md}). The role files are provided verbatim in the released repository (\texttt{prompts/}).

\paragraph{MCP tool server.}
Blender 5.1 runs as a long-lived process; an MCP server exposes the tool-use APIs of \S\ref{sec:tools} (Table~\ref{tab:tools}) over a local TCP port. \texttt{blender\_execute} runs arbitrary bpy Python in the live process; the inspect and preview tools wrap deterministic Python helpers that read scene state without re-rendering. Checkpointing saves the live .blend after every closed stage, so a coder failure on stage $s_{i+1}$ rolls back to the checkpoint written at the close of $s_i$.

\begin{table}[tp]
\centering
\footnotesize
\setlength{\tabcolsep}{3pt}
\renewcommand{\arraystretch}{1.16}

\begin{tabular}{@{}P{0.19\linewidth}P{0.30\linewidth}P{0.46\linewidth}@{}}
\toprule
\textbf{Category} & \textbf{Tools} & \textbf{Purpose} \\
\midrule
\multicolumn{3}{@{}l}{\textbf{\textit{State Observation}}}\\[-1pt]

\skill{Scene Snapshot} &
\textbf{\texttt{blender\_scene\_state}}
&
Structured readout of the live scene: objects, materials, modifier stacks, f-curves, and physics/cache state. \\

\skill{Visual Preview} &
\texttt{blender\_single\_mesh\_preview}\newline
\texttt{blender\_object\_preview}\newline
\texttt{blender\_system\_layout\_preview}\newline
\texttt{blender\_scene\_layout\_preview}
&
Targeted multi-angle preview renders at mesh, object, system, and whole-scene granularity. \\

\skill{Motion Audit} &
\textbf{\texttt{blender\_motion\_sheet\_preview}}
&
Samples a per-actor frame strip with phase boundaries and bake state to surface motion-mechanism failures. \\

\group{State Modification}

\skill{Code Execution} &
\texttt{blender\_execute}
&
Runs bpy code in the live Blender process; fresh namespace per call. \\

\skill{Protocol Tagging} &
\texttt{blender\_tag\_object}
&
Writes the scene-protocol custom properties (\texttt{protocol\_role}, \texttt{is\_object\_root}, \texttt{contact\_with}) that make state checkable by the verifier. \\

\skill{Checkpoint} &
\texttt{blender\_save}
&
Saves the live .blend as the per-stage checkpoint. \\

\skill{Render} &
\texttt{blender\_render}
&
Renders the final frame sequence (or still) for the scene. \\

\group{Knowledge \& Reference}

\skill{API Lookup} &
\texttt{blender\_docs}
&
Queries the auto-derived Blender API knowledge base. \\

\bottomrule
\end{tabular}

\caption{Tool suite exposed to \ours{} agents, grouped by category. Bolded tools are the engine-state inspection tools that supply the mechanism-level evidence the rendered image cannot.}
\label{tab:tools}
\end{table}

\paragraph{Knowledge base.}
\texttt{blender\_docs(query)} is backed by \texttt{knowledge/blender\_docs/}, an auto-generated reference compiled per Blender version by \texttt{src/knowledge/build\_knowledge\_base.py} (it extracts class hierarchies, property signatures, and enum values, and serves curated markdown pages with raw JSON fallback). Consistent with \S\ref{sec:knowledge}, this is the only knowledge source: no hand-curated reference material is shipped. The coder queries it on demand with \texttt{blender\_docs(topic)} for the stage at hand.

\paragraph{Asset Licences.}
Third-party assets used: Blender 5.1 (GPLv2+); Anthropic Claude Opus~4.7 (agent LLM, consumed via the Anthropic API under its Terms of Service); OpenAI GPT-5.5 (evaluation VLM judge, consumed via the OpenAI API under its Terms of Service); BlenderBench~\citep{viga2025} (used for editing evaluation only; license per the BlenderBench repository).

\section{Scene Protocol and Verifier Ruleset}
\label{app:protocol}

This appendix gives the full ruleset for the in-loop verifier introduced in \S\ref{sec:protocol}. It is part of the \ours{} \emph{system} (the deterministic gate the generation loop runs after every stage) and is separate from the external benchmark audit of Appendix~\ref{app:rubric}, which scores any method's output and shares no code with it.

\paragraph{Protocol structure.}
Objects are organised into a three-level collection tree: the scene root (L1); \emph{system} collections (L2, \texttt{protocol\_role}=\texttt{system}) that group logical objects or hold scene-level state, including a mandatory \texttt{ground} and the conventional \texttt{lighting}/\texttt{cameras}/\texttt{physics} collections; and \emph{object} collections (L3, \texttt{protocol\_role}=\texttt{object}), each holding the meshes of one logical object plus exactly one \texttt{Empty} root to which they are parented. Surface contact, absent from Blender's native vocabulary, is declared via the \texttt{contact\_with} custom property and the protocol-graph relations \textsc{SupportedBy}, \textsc{StableAgainst}, \textsc{FixedAttachment}, and \textsc{Aperture}; a pair counts as declared if either side names the other. The verifier accepts the declaration only if the surfaces actually meet (rule V1).

\paragraph{Ruleset.}
The verifier reports \texttt{ok}=false if any \emph{hard} rule fails; warnings are listed separately and do not flip \texttt{ok}. Rules fall in four families: structural (C, G), geometric (V), soft (W), and plan-vs-state (R, T).

\emph{Structural, collection layout (C, cheap pure-data):}
\begin{itemize}\setlength\itemsep{0.1em}
\item \textbf{C1} every L3 sits in $\geq 1$ system L2; \textbf{C2} every protocol L2 sits directly under the scene root; \textbf{C3} every mesh sits in exactly one object L3; \textbf{C4} every L3 has exactly one \texttt{is\_object\_root} Empty; \textbf{C5} all protocol collections are explicitly named (no \texttt{Collection.001} debris); \textbf{C6} a \texttt{ground} system L2 exists with $\geq 1$ L3 inside. All hard fails.
\end{itemize}

\emph{Structural, Empty and parent--child (G, cheap pure-data):}
\begin{itemize}\setlength\itemsep{0.1em}
\item \textbf{G1} every mesh's parent chain ends at its L3's root Empty; \textbf{G2} every root Empty has \texttt{parent=None}; \textbf{G3} every root Empty is in its own L3. All hard fails.
\end{itemize}

\emph{Geometric (V, BVH-based, heavy):}
\begin{itemize}\setlength\itemsep{0.1em}
\item \textbf{V1} every declared contact relation (\texttt{contact\_with} and the protocol-graph relations, cross-object and within-object) must resolve to a nearest-surface distance $\leq \epsilon_m$, measured by BVH on the evaluated meshes. Declaring an attachment never exempts the parts from touching: \emph{declared contact $\Rightarrow$ measured contact} at every level. Hard fail.
\item \textbf{V2} any AABB-overlapping mesh pair with penetration depth above a numerical-noise tolerance must be declared on at least one side, else it is flagged as an undeclared penetration. Hard fail.
\end{itemize}

\emph{Soft (W, warnings):}
\begin{itemize}\setlength\itemsep{0.1em}
\item \textbf{W1} every L3 reachable from a ground L3 via contact edges (promoted to hard fail under \texttt{strict\_grounding}); \textbf{W2} meshes within an L3 form one connected component (skipped for L3s flagged \texttt{allows\_disconnected}); \textbf{W3} no light/camera/force-field inside an object L3; \textbf{W4} scene has $\geq 1$ camera; \textbf{W5} scene has a light or an emissive world.
\end{itemize}

\emph{Plan-vs-state (R, T):}
\begin{itemize}\setlength\itemsep{0.1em}
\item \textbf{R1--R18} re-check the planner's \texttt{protocol\_graph} relations against the realised geometry (e.g.\ a \textsc{Distributed(circle, 4-fold)} layout is actually a 4-fold circular arrangement to tolerance); seven motion-timing rules (the \textbf{T} family; \textbf{T1} and \textbf{T5} are folded into the motion-sheet preview and R15) re-check the realised motion against the planned phases (e.g.\ a \texttt{settle} phase has near-zero velocity in its final frames). The full per-rule catalogue and tolerances are in the released code.
\end{itemize}

Together V1 and V2 catch the geometric failure mode that naive structural checks miss: V1 rejects a declared contact that does not hold (a chair leg floating above the floor), and V2 rejects an undeclared interpenetration (a mesh punching through another), so a scene cannot pass by being merely well-organised while being geometrically broken.

\section{Evaluation Protocol: Engine-State Audit and VLM Judge}
\label{app:rubric}

\bench{} scores each scene on two independent tracks (\S\ref{sec:experiments}): a deterministic engine-state audit that measures \emph{mechanism} correctness from Blender's runtime state, and an itemized VLM judge that measures whether the prompt's content is \emph{visually} delivered. The two are deliberately disjoint (the audit never looks at a pixel, the judge never inspects a modifier) so that a scene which looks right but is built the wrong way (the dominant failure mode of \S\ref{sec:experiments}) scores high on one track and low on the other rather than passing both.

\subsection{Engine-state audit}

The audit runs inside a headless Blender process and reads the generated scene's runtime state directly. It is \emph{system-agnostic}: it inspects bpy data (modifier stacks, physics caches, rigid-body world, collision settings, constraints, force fields, and the per-channel keyframe density on location/rotation) and depends on no \ours{}-specific protocol, so it scores any method's .blend file on identical terms.

\paragraph{Ground truth.}
Each scene carries a hand-authored specification (a YAML file) listing the \emph{expected actors}, each with a \texttt{role} (\texttt{cloth}, \texttt{fluid\_domain}, \texttt{rigid\_active}, \ldots), a set of \texttt{must\_have} predicates, and a set of \texttt{must\_not\_have} predicates; the expected \emph{spatial relations} between actors (\textsc{SupportedBy}, \textsc{Inside}, \textsc{OnTopOf}, \ldots); and scene-level anti-cheat assertions. Actors are resolved to objects in the generated scene by name and by matching hints (AABB size band, topology hint, expected collection role).

\paragraph{Predicate library.}
Predicates are drawn from a typed library of 42 checks, grouped by mechanism:
\begin{itemize}\setlength\itemsep{0.15em}
\item \textbf{Universal (2):} actor is renderable; solver modifiers are enabled in both viewport and render.
\item \textbf{Cloth (5):} CLOTH modifier present; cache baked over the frame range; self-collision; pin vertex group; collision partners carry COLLISION.
\item \textbf{Fluid (9):} FLUID modifier and type (DOMAIN/FLOW/EFFECTOR); a domain exists; required flow count; domain cache baked; dynamic effector on a moving actor; guiding velocity; minimum domain resolution; liquid mesh output.
\item \textbf{Rigid body (9):} rigid-body settings and type (ACTIVE/PASSIVE); collision shape; populated rigid-body world; world cache baked; constraint type and resolved partner; disabled collisions on constraint; collision modifier when interacting with a deformable; positive mass.
\item \textbf{Particle (7):} particle system present; type (EMITTER/HAIR); emission source; cache baked; collision partners; emission from a deformed surface (modifier-stack order); force fields present.
\item \textbf{Soft body (5):} SOFT\_BODY modifier; cache baked; collision partners; goal vertex group; rigid interaction partners carry both rigid-body and collision.
\item \textbf{Anti-cheat (5, \texttt{must\_not\_have}):} solver actor does not carry dense location/rotation keyframes (caps faked animation); cloth/soft do not use shape keys as a motion source; static scenes carry no baked caches, no populated rigid-body world, and no spurious solver modifiers.
\end{itemize}

\paragraph{Aggregates.}
For each actor the audit computes the fraction of its \texttt{must\_have}/\texttt{must\_not\_have} predicates that pass; \textbf{MPR} (Mechanism Pass Rate) is the per-scene mean of these per-actor fractions. \textbf{SPR} (Structural Pass Rate) is the per-scene fraction of declared spatial relations that hold, each checked geometrically by nearest-surface distance via BVH on the evaluated meshes, sampled at start/mid/end frames. Both are reported in Tables~\ref{tab:main}--\ref{tab:ablation}.

\paragraph{SPR cross-system fairness.}
SPR scores spatial relations against the live .blend state, which a naive implementation would let advantage \ours{}: its contact checks expand a logical object to its full set of meshes through the scene protocol's grouping, which baselines do not emit (VIGA produces hundreds of ungrouped primitive meshes). We remove this confound by inferring an object's assembly geometrically when no protocol grouping is present: the 3D connected component of meshes in mutual axis-aligned-bounding-box contact, grown from the matched mesh and stopped at the support surface so it cannot trivially absorb its target. This inference is a strict no-op on protocol-compliant scenes (every \ours{} mesh is already grouped, so \ours{}'s SPR is unchanged) and only relaxes the score for baselines: VIGA's macro SPR rises from $0.62$ to the reported $0.70$, every cell monotonically non-decreasing, while MPR is unaffected by the re-match (a shift under $0.01$). The recovered points concentrate in static scenes where objects were correctly placed but ungrouped (e.g.\ \texttt{static\_L1}, $0.22\to0.74$), whereas scenes whose objects genuinely float or interpenetrate stay low (e.g.\ \texttt{static\_L2}, unchanged at $0.10$, a furniture cluster suspended $\sim$0.8\,m above the floor). Even so, structural scoring across systems with different grouping conventions is hard to make fully fair, so we treat the $+0.19$ SPR gap as supporting evidence and lead with MPR.

\subsection{Itemized VLM judge}

A GPT-5.5 judge receives the user prompt and five frames sampled uniformly from the rendered video at $t \in \{0, 25, 50, 75, 100\}\%$ of the clip (a single still for static scenes); it sees neither the planner's intent nor the audit. Instead of numeric scores, it \emph{enumerates atomic items} from the prompt and returns a binary verdict per item, which yields item-level partial credit and a concrete evidence trail. The five dimensions:
\begin{itemize}\setlength\itemsep{0.15em}
\item \textbf{objects\_present}: one item per object the prompt names; \texttt{present}/\texttt{absent} (a disassembled object that no longer reads as its class counts \texttt{absent}).
\item \textbf{spatial\_relations}: one item per stated relation; \texttt{holds}/\texttt{violated} (floating where support is implied, interpenetration, wrong side).
\item \textbf{actions\_visible} (dynamic only): one item per described motion; \texttt{happened}/\texttt{missing}, judged by comparing the first and last frame; empty for static scenes (the dimension is then excluded).
\item \textbf{visual\_quality}: fixed three-item technical checklist (materials assigned, exposure/lighting, no render artifacts); \texttt{ok}/\texttt{broken}.
\item \textbf{aesthetics}: fixed four-item artistic checklist (material fidelity, colour palette, lighting mood, composition); \texttt{good}/\texttt{poor}.
\end{itemize}
Each dimension's score is the fraction of its items with a positive verdict; the reported VLM score is the mean over the dimensions a scene exercises. Mechanism realism (whether motion is a real simulation or keyframed) is explicitly excluded from the judge's remit, since vision models are unreliable on it~\citep{bansal2024videophy}; the audit covers it instead. The full judge prompt, including the per-item output schema and anchor examples, is provided in the released code repository (\texttt{src/bench/vlm\_judge\_rubric.md}).

\section{Benchmark Details}
\label{app:benchmark}

\paragraph{Composition.}
\bench{} contains 50 scenes (Section~\ref{sec:benchmark}): 45 dynamic scenes across five solver categories (cloth, fluid, rigid body, particle, soft body), each split into three difficulty levels with three prompts per level, plus 5 static scenes (three single-room interiors, two scene-scale layouts). The difficulty axis is defined by mechanism complexity rather than by clip length or raw object count:
\begin{itemize}\setlength\itemsep{0.15em}
\item \textbf{D1 (single actor):} a single solver actor in its default configuration: one cloth draping, one fluid pouring, one stack of rigid bodies settling.
\item \textbf{D2 (within-category):} multiple instances of the category, or internal solver complexity within it: self-collision, pinning and goal vertex groups, rigid-body constraints, force fields, multiple flow sources.
\item \textbf{D3 (cross-category):} cross-category interaction realised in a single shot (a rigid block crushing a soft-body slab, particles emitted from a deforming cloth surface, a fluid effector riding an animated rigid body), each requiring two or more solvers to be configured and to interact correctly.
\end{itemize}
Static scenes carry no solver at all; they exercise object inventory, material assignment, and spatial layout (15--20 named objects per single-room scene), and their anti-cheat predicates assert the absence of any physics state.

\paragraph{Example prompts.}
One prompt per difficulty tier (the canonical ground-truth IDs are \texttt{<category>\_<level>\_<nn>}):
\begin{itemize}\setlength\itemsep{0.2em}
\item \textbf{cloth\_D1\_01} (cloth, single actor): ``A red tablecloth drapes over a small round wooden table in a quiet dining room.''
\item \textbf{rigid\_D2\_01} (rigid body, within-category): twelve wooden dominoes arranged in a chain on a table; the first is tipped and the cascade runs to the end (multiple interacting rigid bodies with a baked rigid-body world).
\item \textbf{soft\_D3\_01} (soft body $\times$ rigid body, cross-category): ``A thick green jelly slab on a wooden board is crushed under a falling heavy stone block, the jelly squashing flat under the block as it settles.''
\item \textbf{static\_D1\_01} (static, single-room): ``A furnished living room interior: a three-seat sofa against the back wall with two cushions, a coffee table on a large rug, a framed picture and a round wall clock, a floor lamp, a tall bookshelf holding rows of books and a small potted plant, a television on a low media console, a side armchair near the window, and a basket of magazines on the floor.''
\end{itemize}

\paragraph{Authoring protocol.}
Each scene is specified by a hand-authored ground-truth YAML (Appendix~\ref{app:rubric}) that fixes the expected actors, their required solver predicates, the spatial-relation graph, and the motion phases, alongside the natural-language prompt. Prompts use domain-specific nouns (no \texttt{ball}/\texttt{object} placeholders) with place, time, and material anchors, and each is validated to exercise its category's solver. The difficulty level is fixed by the authored predicate set (the number of interacting actors and whether the required predicates cross a category boundary) rather than inferred from prompt text.

\paragraph{VIGA run protocol.}
VIGA is run from its open-source release in dynamic-scene mode on the \bench{} prompts. Each scene receives the prompt text only, with no target or reference image, matching \ours{}'s text-only setting. VIGA uses its own generator--verifier loop and Claude Opus~4.7 backend, capped at its native budget of 15 rounds; \ours{} instead runs the per-stage bounded-retry loop of \S\ref{sec:pipeline}, so the two budgets are reported as configured rather than forced equal. Both systems' final \texttt{.blend} files are then scored by the identical external audit and VLM judge of Appendix~\ref{app:rubric}.

\section{BlenderBench Setup}
\label{app:blenderbench}

BlenderBench~\citep{viga2025} is reused unchanged from VIGA for the editing comparison in Section~\ref{sec:experiments}. We summarise its structure and metrics here for transparency, and defer to the VIGA paper and the BlenderBench dataset card for the canonical definitions.

\paragraph{Task structure.}
Each of the 27 tasks contains:
\begin{itemize}\setlength\itemsep{0.2em}
\item a Blender scene .blend file (the ``start'' scene);
\item a $512\times512$ reference render of the post-edit scene (the ``goal'' render);
\item a one- to two-sentence natural-language task description (\texttt{task.txt});
\item the reference Python edit script used by the dataset authors to produce the goal render (\texttt{goal.py}); not exposed to the agent.
\end{itemize}
The agent receives the start .blend file, the goal render, and the task description, and emits Python that, when executed against the start scene, should produce a render approximating the goal.

\paragraph{Difficulty levels.}
The 27 tasks split evenly across three difficulty levels (9 tasks each):
\begin{itemize}\setlength\itemsep{0.2em}
\item \textbf{Level~1: Camera adjustment.} Scene contents and lighting are held fixed; only the camera pose differs between start and goal. Example task description: \emph{``Adjust the camera position so that the viewing angle is consistent with the target image.''}
\item \textbf{Level~2: Multi-step attribute editing.} The camera is held fixed; the agent must change two or more lighting, material, or object-geometry attributes within the same task. Example task description: \emph{``First adjust the room brightness, then adjust the size of the character's belly so that it looks like the target image.''}
\item \textbf{Level~3: Compositional editing.} The same attribute changes as Level~2 plus a Level-1 camera change in the same task. Example task description: \emph{``First adjust the room brightness, then adjust the size of the character's belly so that it looks like the target image. You need to adjust the camera angle so that you can see the object you want to modify.''}
\end{itemize}

\paragraph{Evaluation metrics.}
We report VIGA's three reference-comparing metrics, computed by a re-evaluation pass that mirrors the open-source VIGA implementation~\citep{viga2025}:
\begin{itemize}\setlength\itemsep{0.2em}
\item \textbf{PL$\downarrow$ (photometric loss).} Mean squared error between the agent's final render and the goal render, after both are converted to RGB, normalised to $[0,1]$, and resized to the goal-render resolution. Reported on the paper's $\times 100$ scale.
\item \textbf{N-CLIP$\downarrow$ (CLIP distance).} $(1 - \cos\langle\phi(\text{render}), \phi(\text{goal})\rangle) \times 100$, where $\phi$ is the image embedding from the \texttt{openai/clip-vit-base-patch32} model, the same CLIP variant VIGA uses.
\item \textbf{VLM$\uparrow$ (judge score).} A GPT-4o judge is shown the goal render, the agent's final render, and the task description, and assigns four 0--5 integer scores along VIGA's four criteria (task completion, visual quality, spatial accuracy, detail accuracy) using VIGA's verbatim instruction template. We report the per-task mean of the four scores.
\end{itemize}
For both methods, the final renders produced by the sweep were re-scored end-to-end through this implementation, so the head-to-head means in Table~\ref{tab:blenderbench} are computed against identical metric code rather than against per-method scoring pipelines.

The per-level head-to-head numbers are reported in Table~\ref{tab:blenderbench} in the main text.

\end{document}